\documentclass{bmvc2k}

\title{Estimating small differences \\in car-pose from orbits}

\addauthor{Berkay Kicanaoglu\\Ran Tao\\Arnold W.M. Smeulders}{{b.kicanaoglu, r.tao, a.w.m.smeulders}@uva.nl}{1}

\addinstitution{
 QUVA Lab,\\
 University of Amsterdam,\\
 the Netherlands
}
\runninghead{Kicanaoglu et al.}{Estimating small differences in car-pose from orbits}

\usepackage{amsthm}
\usepackage{amssymb}
\usepackage{rotating}
\usepackage{multirow}
\usepackage{epsfig}
\usepackage{graphicx}
\usepackage{amsmath}

\DeclareMathOperator*{\argmax}{argmax}

\theoremstyle{definition}
\newtheorem{definition}{Definition}[section]
\begin{document}
	
	\maketitle
	
	\begin{abstract}
		
		Distinction among nearby poses and among symmetries of an object is challenging. In this paper, we propose a unified, group-theoretic approach to tackle both. Different from existing works which directly predict absolute pose, our method measures the pose of an object relative to another pose, \emph{i.e.}, the pose difference. The proposed method generates the complete orbit of an object from a single view of the object with respect to the subgroup of $SO(3)$ of rotations around the $z$-axis, and compares the orbit of the object with another orbit using a novel orbit metric to estimate the pose difference. The generated orbit in the latent space records all the differences in pose in the original observational space, and as a result, the method is capable of finding subtle differences in pose. We demonstrate the effectiveness of the proposed method on cars, where identifying the subtle pose differences is vital. 
		
	\end{abstract}
	
	%-------------------------------------------------------------------------
	\section{Introduction}
	\label{sec:intro}
	
	While pose estimation has recently gained substantial progress \cite{Su2015RenderFC, Tulsiani2015ViewpointsAK, Massa2016CraftingAM}, distinction among nearby poses and among distant symmetries  of the object remain a hard problem \cite{RedondoCabrera2016PoseEE}. At the same time, pose estimation is still an important problem. In traffic, where pose determines the future direction of the car, it is even vital.
	
\begin{figure}[tp]
		\centering
		\begin{minipage}{0.25\textwidth}
			\centering
			\includegraphics[width=0.8\textwidth]{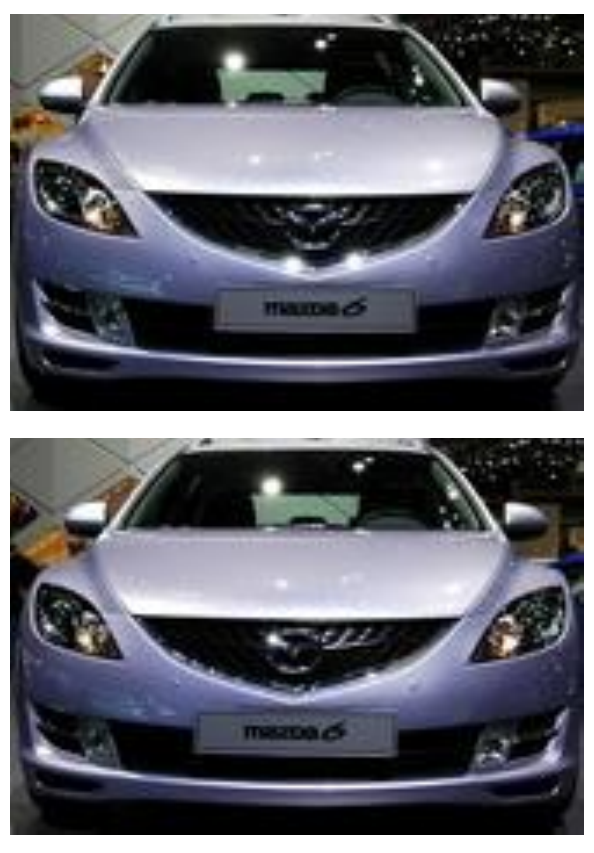} % first figure itself
			\caption{The hardness of distinction among nearby poses.}
			\label{difficulty}
		\end{minipage}\hfill
		\begin{minipage}{0.70\textwidth}
			\centering
			\includegraphics[width=0.9\textwidth]{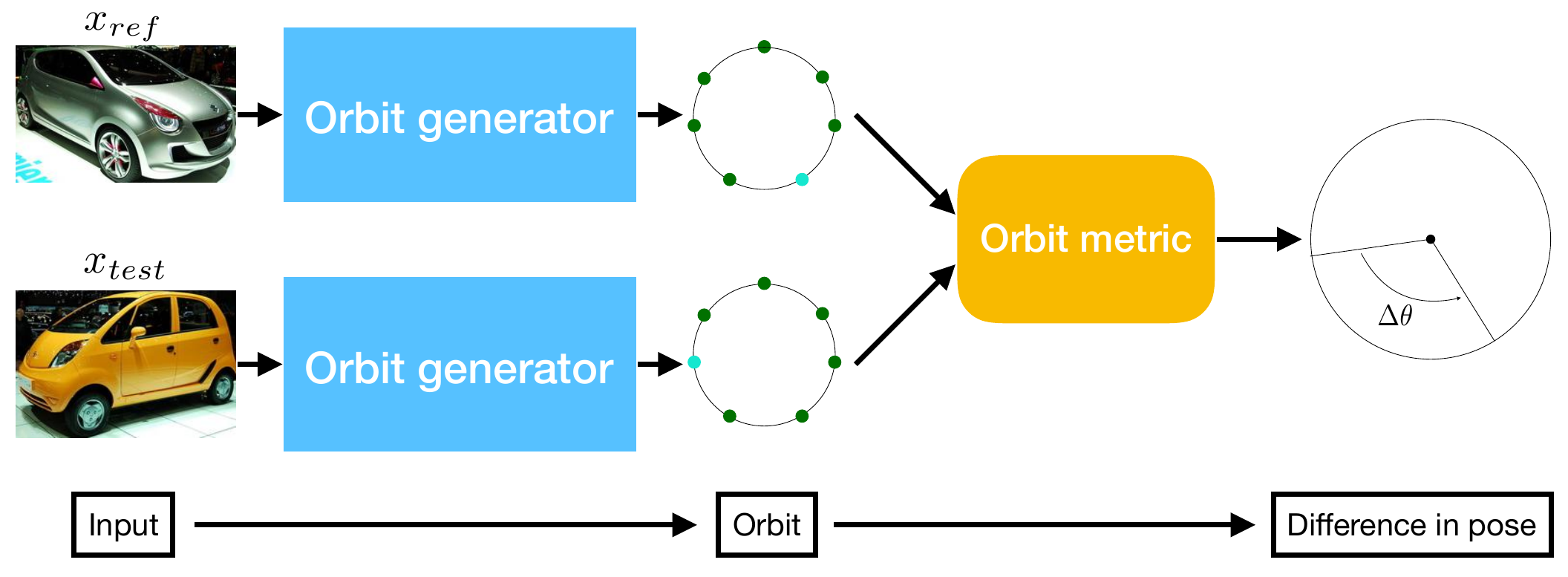} % second figure itself
			\caption{Our approach generates an orbit for each input starting from its initial pose, then compares two orbits to determine the number of steps of rotation necessary to align test object's pose to that of the reference object using the orbit metric.}
			\label{overall-idea}
		\end{minipage}
		% \vspace{-4mm}
	\end{figure}

	Confusion among nearby poses will demonstrate itself at many places but subtly. When the object resembles a tube, no difference in pose can be observed perpendicular to the main tube-axis due to rotational symmetry. When in frontal view or when in view at the round corners of a car, the car may look similar to such a tube. For these types of views, finding pose differences will be hard or very hard, see Figure \ref{difficulty}. In general, the distinction of nearby poses rests on the accumulation of subtle differences distributed over the field of view.
	
	Confusion among symmetries is due to the overall geometry of objects. To discriminate the right side of a car from its left side rests on the detection of sparsely distributed details of difference between the two sides. The distinction of the front of a car from its back requires the detection of sparsely distributed small local differences.
	
	In traffic, exactly these hard conditions are crucial. Distinction among nearby poses is vital to establish the intent of the opposing car to cross before us. And, distinction among symmetries in a split second is vital to determine whether the opposing car is moving towards or away from us. In many cases, for example when following an opposing car in a sequence, changes in pose \emph{relative to its previous pose} are important to detect immediately.
	
	In this work, we take one unified, group-theoretic approach to merge both nearby and symmetry pose distinction into one approach. We construct an \emph{equivariant}, \emph{irreducible} and \emph{interpretable} latent representation to encode the rotation faithfully. Central to our approach is the \emph{orbit} \cite{milneGT}, which is the ordered point set representing all transformed states of a given object with respect to the transformation at hand. In our case, we consider $SO(3)$, \emph{i.e.} the 3D-rotation group.
	
	From this foundation, we propose to generate the orbit per object. We do so in the latent space by learning to rotate them with respect to the subgroup of $SO(3)$ of rotations around the $z$-axis. The orbit records all relative relations in pose between any pair of views. This is relevant as the network is learning how to distinguish among any pair of poses, thus employing all possible information, from nearby, intermediate or distant poses alike. 
	While current methods predict absolute pose \cite{Ghodrati2014Is2I,Su2015RenderFC,Massa2016CraftingAM,Tulsiani2015ViewpointsAK} (\textit{Is this a 50-degree pose?}), our method measures the pose of an object relative to another pose, \emph{i.e.}, the pose difference (\textit{What is the difference between the two poses?}). Absolute pose can be established by gauging it with another known pose. But in many case it is important only to measure subtle relative pose differences.

	% The contributions in this work:
	This paper makes the following contributions. (1) We approach pose estimation from group theory by considering the orbit with appropriate constraints including an orbit-based metric to measure pose difference. (2) A novel network is proposed to generate the orbit of an object from a single view of the object after a learning phase from multiple views with known poses of other objects. (3) We undercut the necessary large amount of data by a learning strategy using synthetic as well as real data.

	In our experiments we will evaluate on the basis of absolute pose, as is common to do, noting that relative pose difference (for example comparing with the previous view) would reveal the qualities of the method more favorably. We evaluate the method on car images as there the two hard conditions are vital: distinction among nearby poses and distinction among symmetries of an object. We observe that among similar network-models we achieve top-performance on these hard but important cases.

	%%%%%%%%%%%%%%%%%%%%%%%%%%%%%%%%%%%%%%%%%%%%%%%%%%%%%%%%%%%
	\section{Related Work}
	Pose estimation has been treated as a regression task in \cite{Fenzi2013ClassGM} from local descriptors of the same patch in different viewpoints. \cite{RedondoCabrera2014AllTN} uses probabilistic regression based on Hough Forests with an uncertainty criterion for continuous pose estimation. Similarly, \cite{Fenzi2015ContinuousPE} uses regressors on Fisher-encoded vectors extracted from spatial cells. \cite{Ghodrati2014Is2I} trains a separate SVM-classifier per viewpoint bin using off-the-shelf CNN-features.  \cite{Massa2016CraftingAM,Su2015RenderFC} adopt data-driven classification by CNNs. In \cite{Su2015RenderFC}, the authors propose a rendering pipeline leveraging the vast amount of CAD-models obtained from \cite{shapenet2015} to generate a large synthetic training set. Moreover, they propose a geometry-aware classification loss to encourage correlation among neighboring views. Similarly, \cite{Massa2016CraftingAM} uses a multi-task CNN pretrained by ImageNet based on AlexNet \cite{krizhevsky2012imagenet} or VGG16 \cite{Simonyan2014VeryDC} for joint detection and pose estimation. These methods predict the absolute pose rather than explicitly considering relations among different views of an object. In contrast, this paper aims to estimate the pose \emph{difference} between views by taking into account the relative relations between any pair of views during learning. 
	
	To resolve ambiguities resulting from symmetries, \cite{Fenzi2014EmbeddingGI} propose to incorporate geometric graph-matching constraints over the keypoint-features. Similarly, \cite{Tulsiani2015ViewpointsAK} combines the merits of global and local representations by jointly learning keypoint prediction and viewpoint estimation using CNNs. The method requires keypoint annotations. In contrast, we also aim to gain robustness to symmetry-confusion but without special annotation. We do so by faithfully embedding the topology of the 3D-rotation group, $SO(3)$, into our pose representation via hallucinating a given object's views from other angles. 
	
	CAD-models have been used for pose estimation. \cite{aubry2014seeing} uses 3D-models and part-detectors to establish correspondences between CAD-models (of chairs) with real images. \cite{Pepik2012Teaching3G} learns to align parts detected by DPM at various viewpoints to a corresponding CAD-model. Similarly, \cite{Lim2013ParsingIO} relies on CAD-models to align parts globally as well as locally in order to improve the pose alignment of objects. \cite{Su2015RenderFC} also uses CAD-models but they render a large amount of 2D-images to learn robust pose classifiers. Similar to the reference, we rely on CAD-data to generate 2D-views for its precise viewpoint annotations and ease of generating sequences of rotating objects. Unlike \cite{Pepik2012Teaching3G, aubry2014seeing, Lim2013ParsingIO}, we do not employ CAD-models for any kind of point or part alignments, but rather use the rendered 2D-images from CAD-models to learn a pose representation describing the orbit of all poses. 
	
	The proposed method generates the orbit of all poses given a single view of an object and then measures the relative distance in azimuth by gauging the orbit of this object against another object's orbit using orbit metric as illustrated in Figure \ref{overall-idea}.
	
	%%%%%%%%%%%%%%%%%%%%%%%%%%%%%%%%%%%%%%%%%%%%%%%%%%%%%%%%%%%%%%%%%
	\section{Background}
	% \subsection{Definitions}
	\theoremstyle{definition}
	
	\begin{definition}{\textbf{(Group)}}	
		A group is a tuple ($G$, $\cdot$) consisting of a set $G$ and its binary operation, $\cdot : G \times G \rightarrow G, ~(g, h) \mapsto g \cdot h$, where $g,h\in G$, satisfying axioms of \emph{associativity}, \emph{closure}, and the existence of an \emph{identity} and an \emph{inverse} element. We use $G$ to denote a group for convenience. 	
	\end{definition}
	
	\noindent
	We deal with a subgroup, $G$, of SO(3) for azimuthal rotations around $z$-axis. The (sub-)group structure is given as: $      G =\left\lbrace e, g ~ |~ e = g_K, ~g_k = \prod_{1}^{k} g\right\rbrace$ where $e,g$ and $K$ denote the identity element, the generator and the order of the group, $\vert G\vert$, respectively. Group element $g_k$ corresponds to $k$- rotations by applying generator $g$, $k$ times. $G$ is characterized by its generator matrix, $g \in \mathbb{R}^{3\times3}$, which acts on  $\mathbb{R}^3$. This subgroup is a \emph{cyclic} group as applying $g$ $\vert G\vert$ times brings the object back to its initial state. For rotations around the $z$-axis, $g$, is:
	\begin{align}
		g = R_z(\theta) = \begin{bmatrix}
			cos(\theta) & -sin(\theta) & 0           \\[0.3em]
			sin(\theta) & cos(\theta)           & 0 \\[0.3em]
			0           &0 &1
		\end{bmatrix},
	\end{align}
	where $\theta \in [0, 2\pi)$ is the degree of rotation.

	\theoremstyle{definition}
	
	\begin{definition}{\textbf{(Orbit)}}
		Let us denote the set, upon which group $(G, ~\cdot)$ acts, with $X$. Then, an \emph{orbit of $x\in X$} undergoing a group transformation, $G$, is the subset of $X$ such that $G \cdot x :=  \left\lbrace g\cdot x ~\vert ~ g \in G \right\rbrace$.	
	\end{definition}

	\noindent
	Informally, the orbit is the set of measurements of an object, $x$, undergoing a group transformation, $G$. 
	In this work, we evenly sample the azimuth into $K=36$ discrete poses such that $\Theta = \left\lbrace k\cdot \Delta\theta ~|~ \Delta\theta =  \frac{2\pi}{K}, k = 0,\dots,K-1\right\rbrace $. $K$ determines $\vert G\vert$ and therefore the generator matrix is parametrized by $\Delta\theta$ such that $g=R_z(2\pi/K)$. In this way, an orbit is an ordered set of $K$ samples on $K$ consecutive poses.

	%%%%%%%%%%%%%%%%%%%%%%%%%%%%%%%%%%%%%%%%%%%%%%%%%%%%%%%%%%%%%%%%%
	\section{Method}

	\subsection{Orbit Generator}
	\label{sec:OrbitGenerator}
	
	\begin{figure*}
		\centering {
			\includegraphics[width=\linewidth]{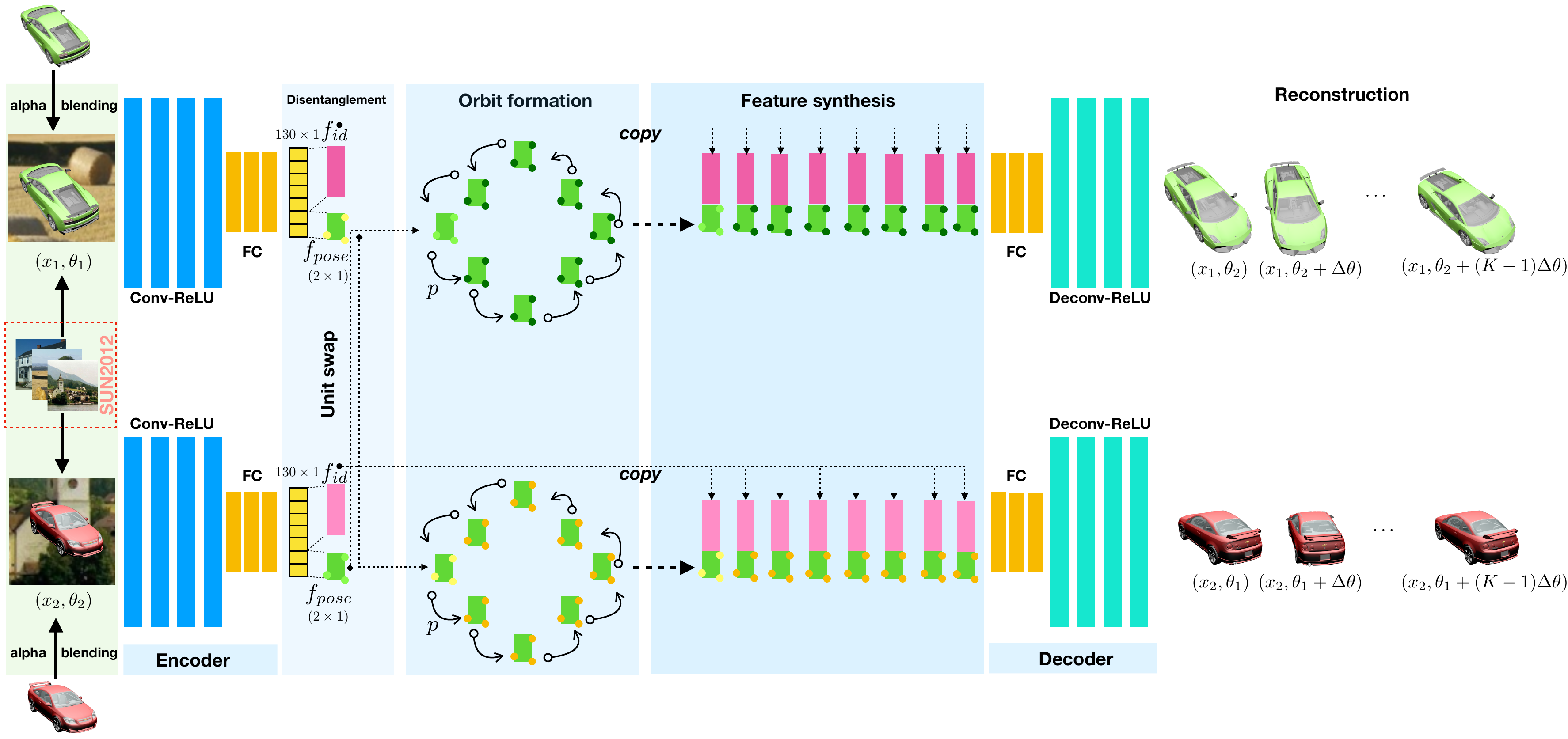}
		}
		% \vspace{-5mm}
		\caption{The proposed two-branch encoder-decoder architecture with shared parameters to learn the orbit generator. After learning, the orbit generator is simply the single branch without the decoder, which takes an input image and generates the full orbit in the latent pose space.} 
		\label{fig:model}
		% \vspace{-3mm}
	\end{figure*}
	
	From a given view, the orbit generator should be capable of \textit(i) inferring the initial pose of the input and \textit{(ii)} constructing the rest of the orbit in latent pose space. We propose a two-branch encoder-decoder architecture with shared parameters to learn the orbit generator (see Figure \ref{fig:model}). To ensure robustness against symmetries, we disentangle two intertwined factors, appearance and 3D-pose, on the final fully-connected layer of the encoder.
	
	The \emph{encoder} subnetwork receives a $64\times 64$ RGB input image and projects it to the latent representation $f_e$ with the map $\mathcal{F}_{encoder}: \mathbb{X}^{64\times 64\times 3} \rightarrow \mathbb{R}^{130}$. It consists of four convolutional layers and three fully connected layers all but the last one followed by a \emph{ReLU}. For the final fully connected layer we opt for \emph{tanh} to make the pose representation cover the range between $-c$ and $c$ symmetrically. As the \emph{ReLU} discards the negative half-plane it is unsuited here. All convolutional layers have a stride of 2 and a receptive field size of $5\times 5$.
	
	For the \emph{disentanglement} we construct a latent pose representation which varies only when the input is transformed by the subgroup $G$. Therefore, we use the irreducible representation to separate the appearance information from the pose information. The first $128$ dimensions of $f_e$ are designated to appearance independent of pose and the final 2 dimensions are designated to pose only. From now on, we refer to the former as $f_{id} ~(\in \mathbb{R}^{128})$ and to the later as $f_{pose} ~(\in \mathbb{R}^2)$.
	
	Our goal is to establish a \emph{homomorphism} between the subgroup $G \subset SO(3)$ defined in $\mathbb{R}^3$ and a group on the latent pose subspace, $(P, ~\ast)$. The homomorphism between $G$ and $P$ implies that $f_{pose}$ will transform in an analogous manner as the action of $G \subset SO(3)$ in $\mathbb{R}^3$. Hence, we have obtained an equivariant representation for the pose. The structure of the group $P$ is similar to that of G, given as $P =\left\lbrace e, p ~ |~ e = p_K, ~p_k = \prod_{1}^{k} p\right\rbrace$. We assume a matrix multiplication for the group operator, $\ast$. We adopt the analytical form of rotation matrix defined in $\mathbb{R}^2$ for its generator $p$ such that the irreducible representation is a block-diagonal matrix of the form
	\begin{align}
		\mathbf{R}(\Delta\theta) =  \begin{bmatrix}
			1 & & &\\
			& \ddots & &\\
			& & 1 	&\\
			& & & p	
		\end{bmatrix} = \begin{bmatrix}
			\mathbf{1} & & &\\
			& & cos(\Delta\theta) & -sin(\Delta\theta)	 \\
			& & sin(\Delta\theta) & cos(\Delta\theta)
		\end{bmatrix}
	\end{align}
	where $\mathbf{1}$ is an identity matrix of size $128\times 128$. Applying $\mathbf{R}(\Delta\theta)$ consecutively leaves $f_{id}$ invariant to rotations due to the identity block but transforms $f_{pose}$. In order to decrease computational redundancy due to large matrix-vector multiplications, i.e. $\mathbf{1} f_{id}$, we use vector slicing to separate $f_{id}$ and $f_{pose}$ and then only apply $p$ on $f_{pose}$ which implies only a $(2\times2) \times (2\times1)$ matrix-vector multiplication. 
	
	Due to the \emph{projection} from 3D to 2D during image capture, the group structure which is observable in 3D is broken. Thus, we cannot establish a homomorphism relying on 2D-images. We impose three constraints on $f_{pose}$ which are derived from the relationships of the group elements in order to get around this problem. Given two inputs $(x_1, \theta_1)$ and $(x_2, \theta_2)$, the encoders compute the corresponding latent pose representations $f_{pose}(\theta_1)$ and $f_{pose}(\theta_2)$. We simultaneously impose three constraints on these latent representations in order to obtain orbits that satisfy the equivalence relations. The first constraint demands that the orbit is a circle over which each pose representation rests, as the group we are dealing with is a cyclic group. We denote it with $\mathcal{L}_{radius}$. The second constraint is derived from the definition of group and orbit. The constraint states that there must be a group action, $p_\star$, that relates any given two elements of the orbit such that $f_{pose}(\theta_2) =  p_\star \ast f_{pose}(\theta_1)$. We refer to it as $\mathcal{L}_{pair}$. And the final constraint is a consequence of the symmetry-relation that orbits must satisfy: given $x,~y \in X$, $y \in G\cdot x \iff x \in G \cdot y$. It implies that $x$ is an element of $y$'s orbit if and only if $y$ is in $x$'s orbit. It yields that any two $f_{pose}$ units can be swapped as they belong to the same orbit. In Table \ref{tab:Constraints}, we provide the loss terms for each of the constraints to be used in the optimization. 
	
	\begin{table}[]
		\centering
		\begin{tabular}{ll}
			\hline
			\multicolumn{1}{c}{\textbf{Constraint type}} & \multicolumn{1}{c}{\textbf{Loss term}}                                                              \\ \hline
			\textit{Radius}                              & $\mathcal{L}_{radius} = \vert c - \vert\vert f_{pose} \vert\vert_2 \vert$                           \\
			\textit{Pair}                        & $\mathcal{L}_{pair} = ||f_{pose}(\theta_2) - (\prod_{1}^{N} p)f_{pose}(\theta_1)||_2$,~ $N = \frac{\theta_2 -\theta_1 \pmod {2\pi}}{\Delta\theta}$ \\
			\textit{Symmetry}                            & Unit swap                                                                                           \\ \hline
		\end{tabular}
		\caption{We impose three constraints on the latent pose representation, $f_{pose}$. Two of them are implemented into the objective function while the last one is implemented into the network architecture. We use $c=0.8$ in all experiments.}
		\label{tab:Constraints}
		% \vspace{-5mm}
	\end{table}
	After the pose swap, \emph{local linear transformations} are applied to compute consecutive pose unit corresponding to the next pose parameter. The next element on the circle can be computed from the previous one by matrix multiplication $f_{pose}(\theta^\prime + \Delta\theta) = p \ast f_{pose}(\theta^\prime)$. We represent the orbit generated by the group $P$ with the set, $\mathcal{X}_p$, \emph{ordered} with respect to the rotations imposed by group elements: $      \mathcal{X}_p=\left\lbrack e\ast f_{pose}, ~p_1\ast f_{pose}, \dots, ~p_{K-1}\ast f_{pose} \right\rbrack   \subset \mathbb{R}^2$. 
	
	The architecture of the \emph{decoder} is mirrored from the encoder where the convolutions are replaced with deconvolutions. The identity unit is copied and concatenated with each and every orbit element in $\mathcal{X}_p$ to give $f = \left\lbrack f_{id} \oplus (e\ast f_{pose}), \dots, ~f_{id} \oplus (p_{K-1}\ast f_{pose})\right\rbrack$. The decoder receives these feature representations and projects them back to the pixel space for reconstruction with $\mathcal{F}_{decoder}$. We denote the output of decoder block with $X_{decoder} = \mathcal{F}_{decoder}[f]$. Then we use the reconstructed image sequences $X^i_{decoder}$ and $X^j_{decoder}$ on both branches given a pair of input images $(x_i, \theta_i)$ and $(x_j, \theta_j)$ to optimize the model with 
	\begin{align*}
		\mathcal{L}_{recon.} = \frac{1}{2 K N_b}\sum_{i,j=1}^{N_{b}} \vert\vert X^i_{decoder} - X^i_{gt}\vert\vert^2_2 + \vert\vert X^j_{decoder} - X^j_{gt}\vert\vert^2_2
	\end{align*}
	where $N_b$, $K$ and $X^i_{gt}$ are the batch size, the order of the group $P$ and groundtruth images depicting the true sequence of the rotations for input image $x_i$.
	
	After learning, the orbit generator is simply the single branch excluding the decoder. It takes an image as input and generates the full orbit in the latent pose space. The orbit in the latent space preserves all information between any views in the original observational space. The orbit representation is compact, a matrix of $K \times 2$ in our case. And, the orbit can easily be visualized to demonstrate where confusion in nearby poses and symmetries (opposite poses) originate.

	\subsection{Orbit Metric}
	To establish whether two poses are equal, one usually measures one pose, $\theta_1$, then the other one, $\theta_2$, and compares the two: $\theta_1 - \theta_2 \pmod {2\pi}$. We argue that a more robust difference between two poses is obtained from comparison of the two complete orbits.
	
	An orbit metric between two orbits, $\mathcal{X}^1_p$ and $\mathcal{X}^2_p$, is defined as:
	\begin{align}
		\mathcal{M}_{\mathcal{X}^{1}_p \leftarrow \mathcal{X}^{2}_p}(\delta) &= \sum_{k=0}^{K-1} <\mathcal{X}^{(1,k)}_p, ~p_\delta\ast\mathcal{X}^{(2,k)}_p> \\
		\mathcal{M}_{\mathcal{X}^{1}_p \leftarrow \mathcal{X}^{2}_p}(-\delta) &= \sum_{k=0}^{K-1} <\mathcal{X}^{(1,k)}_p, ~p^{-1}_\delta\ast\mathcal{X}^{(2,k)}_p>\\
		\Delta\delta &= \min \left\lbrace\argmax_{\delta=0,..,\text{K-1}}\mathcal{M}(\delta), \argmax_{\delta=0,..,\text{K-1}}\mathcal{M}(-\delta)\right\rbrace\label{orbi_metric}
	\end{align}
	Subscript {\footnotesize$\mathcal{X}^{1}_p \leftarrow \mathcal{X}^{2}_p$} denotes that {\footnotesize$\mathcal{X}^{1}_p$} is the reference orbit whereas {\footnotesize$\mathcal{X}^{2}_p$} is the test orbit. We use the notation of {\footnotesize$\mathcal{X}^{(i,k)}_p$} to denote the $k$th element in {\footnotesize$\mathcal{X}^{i}_p$}. Also, the second term inside {\footnotesize$<\cdot, ~\cdot>$} corresponds to a cyclic shift in {\footnotesize$\mathcal{X}^{2}_p$} implying that all the elements are rotated by $\delta\Delta\theta$ or $-\delta\Delta\theta$ depending on the way orbits are compared. 
	
	Rather than the orbit metric (\ref{orbi_metric}), we prefer to use the directed orbit comparison, $\mathcal{M}$:
	\begin{align}
		\Delta\delta &= \argmax_{\delta=0,..,\text{K-1}}\mathcal{M}(\delta)
	\end{align}
	Ideally, the directed orbit comparison produces a sinusoidal signal varying with the shift parameter, $\delta$. We estimate the pose of the test image by:
	\begin{align}
		\theta_{test} &= \theta_{ref} + \Delta\delta\Delta\theta \pmod {2\pi}
	\end{align}
	
	The directed orbit comparison satisfies the \emph{non-negativity}, \emph{identity of indiscernibles} and \emph{triangular inequality} properties of a metric but fails the \emph{symmetry} property. The directed comparision enables us to encode the direction to the left or to the right of the relative pose.
	
	\subsection{Training}
	In the first stage of learning the orbit generator, we need precise orientation data in order to arrive at precise orientation estimates. In practice, for cars (and many other objects) this implies a choice for synthetic training data as only synthetic data can guarantee accurate orientation annotations. For this reason, we acquire 1350 car models from ShapeNet \cite{shapenet2015} and render each one at $10^\circ$ intervals of azimuth and elevation within $[-10^\circ, 50^\circ]$. %over $[0^\circ, 360^\circ)$ and $[-10^\circ, 50^\circ]$, respectively. 
	
	In the second stage, we infuse real data into the synthetic data to compensate for the lack of reality in the synthetic data. Since established benchmark datasets such as PASCAL3D+ do not contain images of the same object in different poses, we use real data only for imposing latent space constraints. We use the RMSProp optimizer with an initial learning rate of $10^{-4}$ which decays exponentially with $\gamma=0.95$ every 10k iterations.
	
	For data preprocessing,
	we use alpha-blending to embed rendered images onto randomly cropped backgrounds from the SUN2012 dataset \cite{Xiao2010SUNDL} on the fly. Alpha blending is only applied to the input images excluding the target images used in reconstruction. For real images, we resize the detection boxes such that the longest dimension is 64 pixels and zero pad the remaining parts. %We normalize all images to $[0,1]$.
	
	The \emph{optimization objective} combines latent space constraints with the mean-squared reconstruction error :
	\begin{align}
		\mathcal{L}_{total} = \beta_1 \mathcal{L}_{recon.} + \beta_2 \mathcal{L}_{radius} + \beta_3\mathcal{L}_{pair}
	\end{align}
	We take $(\beta_1, ~\beta_2, ~\beta_3) = (100, ~1, ~3)$ during the first stage of optimization and $(\beta_1, ~\beta_2, ~\beta_3)  = (100, ~1, ~5)$ from then on.\\
	\subsection{Infer absolute pose from relative pose difference} 	
	Our method measures the relative pose difference between two views. Absolute pose of a view can be inferred by gauging it with another known pose. In this work, we simply use a synthetic image with its pose label $\theta_{ref}$ as the gauging example.

	%%%%%%%%%%%%%%%%%%%%%%%%%%%%%%%%%%%%%%%%%%%%%%%%%%%%%%%
	\section{Experiments}
	
	\subsection{Datasets}

	For evaluation, we use EPFL Car Dataset \cite{zuysal2009PoseEF} and PASCAL3D+ \cite{Xiang2014BeyondPA}. 
	
	\textbf{EPFL Cars Dataset  \cite{zuysal2009PoseEF}.} It contains varying-length sequences of 20 cars on a rotating stage. The dataset does not have exact labels for pose. Therefore, we follow the standard procedure \cite{zuysal2009PoseEF} and produce approximate labels using the timestamps provided, assuming a constant angular velocity for the rotating platform. Following \cite{Ghodrati2014Is2I}, we use first 10 cars for training and the remaining 10 for testing.% our joint identity and pose recognition approach.}
	
	\textbf{PASCAL3D+ \cite{Xiang2014BeyondPA}.} This dataset is widely used to evaluate object detection and pose estimation tasks. It contains 12 rigid categories from PASCAL VOC 2012 \cite{Everingham15} with 3D annotations. We use the train sets of Pascal3D+ and ImageNet cars to train our model and use the Pascal3D+ validation set for testing using RCNN detections\footnote{We use the RCNN detections provided by~\cite{Su2015RenderFC}.}.
	
	Following standard protocol, the performance on EPFL is measured using \textit{accuracy-36} (36 bins)~\cite{Ghodrati2014Is2I} while the performance on PASCAL3D+ is measured using \textit{AVP-24} (24 bins) \cite{Tulsiani2015ViewpointsAK}.
	\subsection{Results}
	% In this section, we provide with the comparsions to state-of-the-art methods in public benchmarks. 
	
	Our method is designed for measuring relative pose difference. However, in order to compare with state-of-the-art in pose estimation, here we evaluate on the basis of absolute pose.

	\begin{table}[!htb]
		\begin{tabular}{lc}
			\hline
			\multicolumn{1}{c}{Method}                          & accuracy-36 \\ \hline
			3D2PM-C \cite{Pepik20123D2PM3}     & 52.1\%             \\
			3D2PM-D \cite{Pepik20123D2PM3}     & 45.8\%             \\
			Fisher+spm \cite{Ghodrati2014Is2I}    & 51.8\%             \\
			Decaf \cite{Ghodrati2014Is2I}             & 45.9\%             \\
			This paper                                          & 54.0\%             \\ \hline
		\end{tabular}
		\quad
		\quad
		\quad
		\begin{tabular}{llc}
			\hline
			& \multicolumn{1}{c}{Method}                                               & AVP-24                     \\ \hline
			\multirow{2}{*}{\rotatebox[origin=c]{90}{DPM}} & \cite{Xiang2014BeyondPA}                                & 13.7\%                     \\
			& DPM-VOC+VP\cite{Pepik2012Teaching3G}                    & \multicolumn{1}{l}{~24.6\%} \\ \hline
			\multirow{3}{*}{\rotatebox[origin=c]{90}{Shallow}} & \cite{Ghodrati2014Is2I}                                 & 15.9\%                     \\
			& Render For CNN \cite{Su2015RenderFC}                    & 25.5\%                     \\
			& This paper                                                               & 28.3\%                     \\ \hline
			\multirow{2}{*}{\rotatebox[origin=c]{90}{Deep}} & Viewpoints \& Keypoints \cite{Tulsiani2015ViewpointsAK} & 40.0\%                     \\
			& Crafting MT-CNN \cite{Massa2016CraftingAM}              & 44.2\%                     \\ \hline
		\end{tabular}
		\caption{Comparison with the state-of-the-art methods for pose recognition on EPFL (\emph{left}) and PASCAL3D+ cars (\emph{right}). }
		\label{pose-estimation-results}
		% \vspace{-4mm}
	\end{table}

	The results are summarized in Table \ref{pose-estimation-results}. On EPFL, our method achieves state-of-the-art result, surpassing DPM-based \cite{Pepik20123D2PM3} and \cite{Ghodrati2014Is2I} which relies on global CNN activations and separate classifiers. On PASCAL3D+ cars, our method compares favorably against DPM-based methods \cite{Xiang2014BeyondPA,Pepik2012Teaching3G} and deep learning methods~\cite{Ghodrati2014Is2I,Su2015RenderFC} with similar network capacity to ours. 

	\textbf{Error modes.} In the light of the study of \cite{RedondoCabrera2016PoseEE}, we investigate the error modes of our approach and compare against \cite{Xiang2014BeyondPA,Tulsiani2015ViewpointsAK,RedondoCabrera2015BecauseBD,Pepik2012Teaching3G}. The study classifies pose errors into three types: \emph{nearby} ($15^\circ\leq\text{err}\leq30^\circ$), \emph{opposite} ($\text{err}>165^\circ$) and \emph{others} ($30^\circ<\text{err}\leq165^\circ$). We present the results in Table \ref{error-analysis}. Our method achieves the lowest nearby-view error rate. In \emph{opposite} case, %our method outperforms DPM and Hough Forest based approaches. 
	\cite{Tulsiani2015ViewpointsAK} works the best, benefiting from the joint learning with keypoint estimations. Our method does not use keypoints as it requires additional keypoint annotations for learning, and still achieves a relatively low error rate in \textit{opposite} case, better than DPM and Hough Forest based approaches. To conclude, the proposed method which takes into account all the relative relations between any pair of views is advantageous in handling nearby and opposite poses. 

	\begin{table}[ht]
		\centering
		\resizebox{\textwidth}{!}{%
			\begin{tabular}{lccl}
				\hline
				\multicolumn{1}{c}{Method} & nearby ($15^\circ<\text{err}<30^\circ$) & opposite ($\text{err}>165^\circ$) & \multicolumn{1}{c}{others} \\ \hline
				VDPM \cite{Xiang2014BeyondPA} & 13.7\% & 16.55\% & 30\% \\
				DPM-VOC+VP\cite{Pepik2012Teaching3G} & 13\% & 12\% & 20\% \\
				BHF \cite{RedondoCabrera2015BecauseBD} & 14\% & 11\% & 54\% \\
				Viewpoints \& Keypoints \cite{Tulsiani2015ViewpointsAK} & 12\% & 5\% & 18\% \\
				This paper & 9.6\% & 8.9\% & 35\% \\ \hline
			\end{tabular}%
		}
		\caption{Comparison with the state-of-the-art in terms of error modes.} 
		\label{error-analysis}
		% \vspace{-6mm}
	\end{table}
	
	% \subsection{Orbits visualizations}
	
\begin{figure}[htp]	
		\centering
		\includegraphics[width=0.95\textwidth]{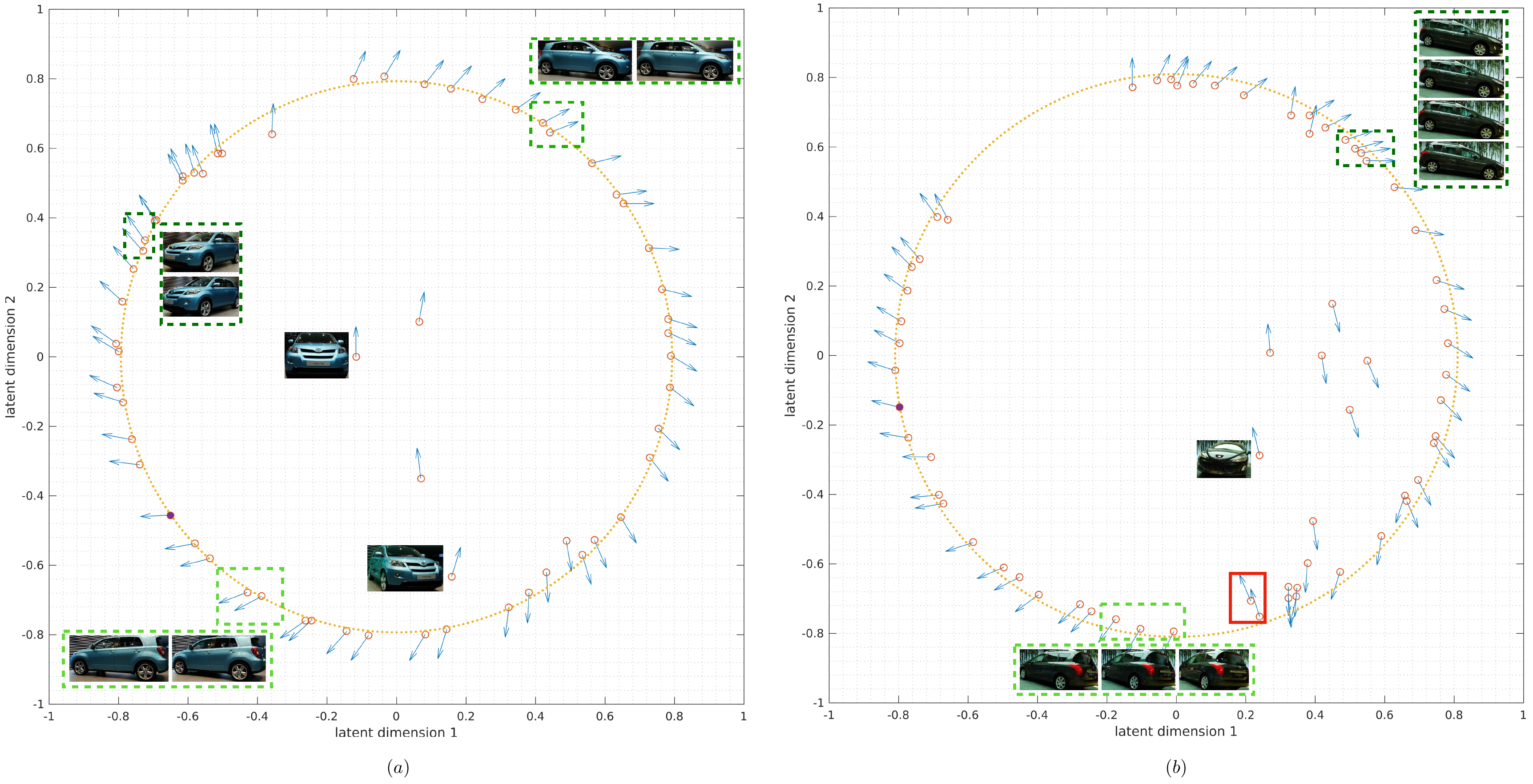}		
		\caption{Visualizations of two orbits.}
		\label{fig:orbits-epfl}
	\end{figure}

	\textbf{Orbits visualizations.} Figure~\ref{fig:orbits-epfl} visualizes the true orbits which is the collection of $f_{pose}$ representations given each view. Dashed circle depicts the trace of a generated orbit starting from the filled initial view. Arrows indicate the ground-truth. Note that the arrows are not meant to show the absolute object orientation. They are used to illustrate the continuously changing pose of a car on a \textit{anticlockwise} rotating stage. 
	As shown in the figure, pose differences in the original observational space are properly captured in the orbits in the latent space thanks to equivariance. We can also see that some opposite views are confused, such as the two views in (b) highlighted by the red box. 
	
	\section{Conclusion}	
	This paper proposes a new method for predicting pose difference by comparing orbits with a metric tailored for it. The method is capable of finding subtle differences in pose. Subtle differences in pose express intent which is important. We have demonstrated the effectiveness of the method on cars, where identifying small pose changes is hard due to their convexity but vital. For similar type of networks, we achieve the best performance.

	% \newpage
	\bibliography{egbib}
\end{document}